\documentclass[conference]{IEEEtran}
\IEEEoverridecommandlockouts
\usepackage{cite}
\usepackage{url}
\usepackage{amsmath,amssymb,amsfonts}
\usepackage{algorithmic}
\usepackage{graphicx}
\usepackage{booktabs}
\usepackage{textcomp}
\usepackage{xcolor}
\usepackage{balance}
\def\BibTeX{{\rm B\kern-.05em{\sc i\kern-.025em b}\kern-.08em
    T\kern-.1667em\lower.7ex\hbox{E}\kern-.125emX}}
\begin{document}

\title{On Estimating the Training Cost of Conversational Recommendation Systems}

\author{\IEEEauthorblockN{Stefanos Antaris}
\IEEEauthorblockA{KTH Royal Institute of Technology \\
Hive Streaming AB \\
Sweden \\
antaris@kth.se}
\and
\IEEEauthorblockN{Dimitrios Rafailidis}
\IEEEauthorblockA{Maastricht University \\
Netherlands\\
dimitrios.rafailidis@maastrichtuniversity.nl}
\and
\IEEEauthorblockN{Mohammad Aliannejadi}
\IEEEauthorblockA{University of Amsterdam \\
Netherlands\\
m.aliannejadi@uva.nl}
}

\maketitle

\begin{abstract}
Conversational recommendation systems have recently gain a lot of attention, as users can continuously interact with the system over multiple conversational turns. However, conversational recommendation systems are based on complex neural architectures, thus the training cost of such models is high. To shed light on the high computational training time of state-of-the art conversational models, we examine five representative strategies and demonstrate this issue. Furthermore, we discuss possible ways to cope with the high training cost following knowledge distillation strategies, where we detail the key challenges to reduce the online inference time of the high number of model parameters in conversational recommendation systems.  
\end{abstract}


\section{Introduction}

Matrix factorization, factorization machines, and learning to rank models have been widely used to predict user preferences and generate recommendations accordingly. The main idea follows the principles of collaborative filtering to learn from users' history records. In conventional recommendation systems, users interact in a straightforward manner, that is, having captured user preferences, the system produces a recommendation list in a single turn~\cite{ThompsonGL04}.
Nowadays, the proliferation of virtual assistants and chatbot technologies opened a new research direction in recommender systems~\cite{Jan20,abs-1901-00431,RafailidisM19}. Conversational recommender systems progressively learn about user preferences over a dialogue, where users provide and refine their feedback online in an interactive way~\cite{Christakopoulou16, Lei0MWHKC20, Jan20, MahmoodR09, SunZ18, ThompsonGL04}. With the advent of deep learning strategies in Natural Language Processing, various neural architectures have been introduced for conversational recommendation, to capture user preferences on items over a dialogue~\cite{Bi19, Kumar16, LiKSMCP18, ZhangCA0C18, ZhangGGGLBD18}. However, deep learning strategies for conversational recommendation design complex neural architectures such as memory networks, recurrent neural networks, and convolutional neural networks. Thus, training these models and learning their parameters leads to high computational costs, which does not necessarily pay off in terms of high recommendation accuracy~\cite{DacremaCJ19}.

High training computational costs  has always been an issue in neural architectures in recommendation systems. Researchers have justified high training costs for effective and efficient outcomes during the test phase. However, the sizes of neural architectures in terms of model parameters significantly increase, making the recommendation mechanisms have an online latency when inferring the test data of user preferences~\cite{BaC14,LeeCLS19, TangW18}. Therefore, a pressing challenge resides on how to estimate the training cost and design a learning strategy, to reduce the latency of online preferences over a dialogue in recommendation systems.

To shed light on the shortcomings of state-of-the-art conversational recommendation strategies, in this paper we make the following contributions:
\begin{itemize}
    \item We first review related studies in conversational recommendation systems.
    \item Then, we detail five representative strategies for conversational recommendation systems. Our analysis is based on the publicly available implementations, provided by the authors of the original work.
    \item In our experiments with four publicly available datasets we show the high number of parameters that is required to train the examined models.
    \item Finally, we discuss our experimental results and provide the key challenges to combat the high online inference time of the model parameters in conversational recommendation systems.
\end{itemize}

\section{Conversational Recommendation Systems}\label{sec:rel}
The Adaptive Place Advisor system~\cite{ThompsonGL04} was among the first attempts to generate conversational recommendations. The system first captures the long-term user preferences over dialogues, which are then used for future conversations to progressively narrow the users' options via a constraint-based recommendation model. 
Mahmood et al.~\cite{MahmoodR09} propose an adaptive model to determine the optimal conversational policy based on a reinforcement learning strategy. Christakopoulou et al.~\cite{Christakopoulou16} introduce a preference elicitation framework based on a probabilistic latent factor model to ask questions to cold-start users so as to learn their preferences quickly. 

Various neural architectures have been introduced for conversation recommendation. For example, Kumar et al.~\cite{Kumar16} present a neural architecture of a dynamic memory network to process input sequences and questions, as well as generate relevant answers. Questions initiate an iterative attention mechanism to condition its attention on the inputs and the result of previous iterations, and then these results are fed into a hierarchical recurrent sequence model to generate recommendations. Zhang et al.~\cite{ZhangCA0C18} adapt a Multi-Memory Network architecture~\cite{SukhbaatarSWF15} and formulate a joint objective function to learn the query representations and produce recommendations over multiple conversational turns. The query module in the neural architecture learns how to ask aspect-based questions in the right order via a ranking loss function so as to capture the user preferences, whereas the search/recommendation module produces the results in a recommendation list. Li et al.~\cite{LiKSMCP18} design a neural architecture of hierarchical recurrent neural networks and autoencoders for movie recommendation. After each dialogue, the hierarchical recurrent neural network's role is to capture/classify the user sentiment/opinion on the movies that were discussed over the dialogue. Then, an autoencoder model is provided with the user's opinion on movies to generate recommendations. Bi et al.~\cite{Bi19} propose a conversational model by exploiting users' negative feedback. Users' preferences are first captured by collecting feedback on certain properties: the aspect-value pairs of the non-relevant items. Then, both positive and negative feedback are exploited based on an aspect-value likelihood model. The Adversarial Information Maximization model~\cite{ZhangGGGLBD18} performs adversarial training to boost diversity in the system's responses to users over the dialogue and then maximizes the variational information maximization objective of InfoGAN~\cite{ChenCDHSSA16} to produce informative responses when generating recommendations. Lei et al.~\cite{Lei0MWHKC20} propose a three-stage Estimation-Action-Reflection process for conversational recommendations. The Estimation module first tries to predict users' preferences on items or on items' attributes via a factorization machine model. The Action module learns a dialogue policy to determine whether to ask attributes of items or directly recommend items via a policy network, and the Reflection module updates the model accordingly when a user rejects the recommendations. 

However, baseline conversational models train their neural models on a huge number of parameters. As a consequence, the vast amount of parameters negatively affects the model performance by having an online inference latency when producing recommendations on the test data of user preferences~\cite{BaC14, LeeCLS19}.

\section{Examined Models}\label{sec:mod}

In a mixed-initiative conversational recommendation setting, the system assumes that after the user initiates a recommendation session, it may take the initiative by asking questions~\cite{DBLP:conf/chiir/RadlinskiC17}. The questions can help the recommendation mechanism understand the user's context and elicit the preferences accordingly~\cite{ZhangCA0C18, radlinski-etal-2019-coached}. There are multiple assumptions about on how a conversation is initiated. Some studies assume that the user initiates a session by asking for recommendation in the form of a question~\cite{ZhangCA0C18,Bi19}. In contrast, another line of research assumes that the system would initiate a session in order to elicit the user's preference~\cite{radlinski-etal-2019-coached}. In this work, we focus on the assumption that the user initiates a conversational recommendation session, as there has been more work performed in this line of research. 

Let $\mathbf{U} = \{u_1,u_2,\dots,u_M\}$ be the set of $M$ users, and $\mathbf{I} = \{\rho_1, \rho_2, \dots, \rho_N\}$ be the set of $N$ items. Following the relevant work~\cite{ZhangCA0C18,Bi19}, we assume that each item comes with a conversation about the item's attributes. The conversation is in the form of questions asked by the system, followed by the user's answer~\cite{ZhangCA0C18,DBLP:conf/sigir/AliannejadiZCC19}. Therefore, a set of question and answers $\mathbf{Q}_i = \{q_1a_1, q_2a_2, \dots, q_{K_i}a_{K_i}\}$ accompanies each item $\rho_i$, containing $K_i$ question-answer pairs, with $i=1,\dots,N$. Furthermore, we denote the user's initial request by $q_0$. Each question-answer pair constitutes a conversation turn $\mathbf{t}$. A recommendation system should ask some questions before it returns its recommendations to the user. \emph{The goal of a conversation recommendation mechanism is to suggest a list of items in the minimum number of conversational turns.}

In the remainder of the Section, we detail the five examined models used in our experimental evaluation to estimate their training costs.

\subsection{Dynamic Memory Networks (DMN)}
DMN~\cite{Kumar16} consists of the input, question, episodic memory, and answer modules. Initially, the input module produces word representations of each item $\rho_i$ via a gated recurrent network (GRU) from raw text inputs. In our case, the initial request $q_0$. Similarly, over the $t$-th conversational turn, the question module generates the respective word representations of a question $q_t$, where the word representations of the input and question modules are stored in a shared matrix. Then, the episodic memory module initiates an iterative process to select which part of the inputs and questions to focus on via an attention mechanism with a gating function. The state of the representations of the $x$-th iteration is preserved in an internal memory $m_x$. In each iteration of the episodic memory module, the representation $m_x$ is updated based on the representations of the question $q_t$ and the computed representations $m_{x-1}$ of the previous iteration. The final calculated representations of the episodic memory module are fed into another GRU to generate answers and provide users with recommendations. DMN is trained by formulating a minimization problem with a cross-entropy loss function of the question/answer sequence $\mathbf{Q}_i$. Note that DMN is a non-personalized conversational model without considering users' representations/preferences. We compute the training parameters using the publicly available implementation\footnote{\url{https://github.com/DongjunLee/dmn-tensorflow}}.

\subsection{Personalized Multi-Memory Networks (PMMN \& MMN)}

Following the neural architecture of DMN, the PMNN model~\cite{ZhangCA0C18} defines two modules, that is, the question and search/recommendation modules. The goals of the two modules are to compute the respective question and item representations over the conversational turns. There are two fundamental differences between DMN and PMMN. (i) PMMN tries to predict the recommended items in the next conversational turns, as DMN does. In the question module, PMMN attempts to predict the question to ask in the next conversational turn. (ii) In addition, PMMN works in a personalized manner. This is achieved by designing a two-layer fully connected network to embed the user representations into the DMN's calculated representations of the episodic memory module. In the PMMN model, a joint loss function is defined with respect to both the query and item representations, and then the model is optimized via stochastic gradient descent. A non-personalized version is also studied, namely MMN, omitting users' representations when training the model. In our experiments, we used the publicly available implementation\footnote{\url{http://yongfeng.me/attach/conversation.zip}}, provided by the authors.

\subsection{Adversarial Information Maximization (AIM)} 

AIM~\cite{ZhangGGGLBD18} is an adversarial learning model for conversational recommendations, built based on the neural architecture of InfoGAN~\cite{ChenCDHSSA16}. According to a dual adversarial objective, two objectives are designed, that is, the answer-to-question (forward) and question-to-answer (backward) objectives. To perform adversarial training, a discriminator is used to differentiate synthetic question-answer pairs from ground-truth question-answer pairs $(q_xa_x)$ of set $\mathbf{Q}_i$, for an item $\rho_i$. Note that the generator produces the synthetic pairs with an architecture of convolutional neural network (CNN) and long-short term network (LSTM) for encoding the word representations. Adversarial training is performed (i) by boosting the diversity of the generated answers over a maximum likelihood learning strategy, and (ii) by maximizing the informativeness of the generated answers based on the variational information maximization objective of InfoGAN. AIM ignores users' personalized selections and computes only the items' representations during the adversarial training, so as to produce recommendations in a conversational turn. We estimate the training cost of AIM based on the publicly available source code\footnote{\url{https://github.com/dreasysnail/converse{\_}GAN}}.

\subsection{Sentimental Analysis via Recurrent Neural Networks and Personalized AutoEncoder Recommendation (sRNN-AE)}

sRNN-AE~\cite{LiKSMCP18} is composed of two modules, a hierarchical recurrent network for performing sentiment analysis to classify users' opinions on items, and an autoencoder strategy to produce personalized recommendations in a conversational turn. The sentiment analysis via the recurrent neural network exploits the question-answer pairs $(q_xa_x)$. It computes the word representations to predict whether the user $u_j$ liked an item $\rho_i$, considering each item $\rho_i$ mentioned so far in the dialogue. These predictions are fed into the autoencoder architecture. In particular, the personalized recommendations are generated by the user-based autoencoder for collaborative filtering (U-Autorec), presented in~\cite{SedhainMSX15}, a model to predict ratings for users not observed in the training set, that is, user ratings on items not discussed so far over the conversational turns. To calculate the training parameters, in our experiments, we used sRNN-AE's publicly available implementation\footnote{\url{https://github.com/RaymondLi0/conversational-recommendations}}.

\section{Experiments}\label{sec:exp}
\subsection{Evaluation Datasets}
We report the experiments on the four publicly available datasets extracted from the Amazon product search dataset~\cite{DBLP:conf/kdd/McAuleyPL15}. The datasets are collected from various categories of items on the Amazon marketplace, namely, \textit{Electronics}, \textit{CDs \& Vinyl}, \textit{Kindle Store}, and \textit{Cell Phones}. We adopt the aspect-value pairs extracted and released by \cite{ZhangCA0C18} and consider them as question-answer pairs. Table~\ref{tab:ds} lists some of the details of the datasets. More details about the creation of the dataset can be found in~\cite{ZhangCA0C18}. Notice that even though the \textit{Electronics} dataset contains more users than the \textit{CDs \& Vynil} dataset, it features less conversational turns. This is the result of more reviews per item in the \textit{CDs \& Vynil} dataset, leading to more aspect-values extracted for each user-item pair.

\begin{table}[h]
    \centering
    \caption{Evaluation datasets}
    \label{tab:ds}
    \begin{tabular}{lccccc}
                            & \textbf{Users}    & \textbf{Items}    & \textbf{Aspects} & \textbf{Values} & \textbf{Conv.~Turns}  \\\hline
        \midrule
         Electronics        &  142,421          & 53,278     & 408 & 493       & 475,020               \\
         CDs \& Vinyl       &  64,847           & 60,405     & 442 & 724       & 659,737               \\
         Kindle Store       &  56,847           & 53,907     & 140 & 338       & 367,159               \\
         Cell Phones        &  21,615           & 9,292      & 276 & 395       & 68,709                \\
        \hline
    \end{tabular}
\end{table}

\subsection{Estimation of Models' Parameters}

For all the examined models presented in Section~\ref{sec:mod}, we preserved the hyperparameters, e.g., number of hidden layers, embedding size, and so on, as they are in the publicly available implementations. Table \ref{tab:model_size} presents the model size of each examined conversational strategy, that is, the number of parameters that need to be trained for each dataset. The model sizes differ, as they examined conversational strategies are based on different neural architectures. For all the four evaluation datasets, the baselines need to learn 0.9-39.1M parameters. DMN has the least number of parameters, as it uses a more simplified neural architecture than the other strategies. The most complex neural architecture of sRNNN-AE requires to train 10.8-39.1M parameters. 

\begin{table}[h]
    \centering
    \caption{Model parameters of the examined strategies}
    \resizebox{.5\textwidth}{!}{ 
    \begin{tabular}{lccccc} \\
    
                        & \textbf{DMN}  & \textbf{MMN}  & \textbf{PMMN} & \textbf{AIM}  & \textbf{sRNN-AE}  \\\hline
        \midrule
        Electronics     & 8,177,095     & 30,979,543    & 31,002,343    & 35,394,441    & 39,168,003        \\
        CDs \& Vinyl    & 5,896,327     & 26,424,743    & 26,445,843    & 28,098,243    & 31,357,475        \\
        Kindle Store    & 5,262,151     & 12,334,643    & 12,351,843    & 15,235,847    & 18,564,869        \\
        Cell Phones     & 979,687       & 5,164,643     & 5,190,343     & 7,269,435     & 10,841,169        \\
        \hline
    \end{tabular}
    }
    \label{tab:model_size}
\end{table}

\subsection{Discussion}
Based on our analysis, it is clear that state-of-the-art models have a high training cost in conversational recommendation systems. The high computational cost to train complex neural architectures does also exist in other research domains.
This has led to an increasing interest in \textit{knowledge distillation}~\cite{DBLP:conf/kdd/BucilaCN06} models that can achieve comparable performance, but with much fewer model parameters. In practice, this means that a student model's shallow network can learn to imitate a deep network of a teacher model, leading to a similar performance of the student and teacher models. In contrast, the student model can significantly decrease the number of model parameters and online latency~\cite{BaC14}. 
Knowledge distillation (aka. knowledge transfer) has been investigated in Information Retrieval~\cite{DBLP:journals/corr/abs-1904-09636}, Natural Language Processing~\cite{DBLP:journals/corr/abs-1910-01108}, Reinforcement Learning~\cite{DBLP:journals/corr/RusuCGDKPMKH15} and compression~\cite{DBLP:conf/iclr/PolinoPA18}. In the context of the classification problem, knowledge distillation models have followed the learning strategy of teacher and student models on the training and test data, respectively~\cite{HeoLY019a, HintonVD15, YimJBK17, WangZSQ18}. The larger teacher model is first trained on the training data and then used to supervise the smaller student model's learning using its output as soft labels. This is achieved by designing a distillation loss function that exploits the teacher model's trained parameters to guide the student model's learning. Recently, a few attempts have been made to design knowledge distillation models in recommendation systems, following the collaborative filtering strategy and focusing on the ranking performance in the top-n recommendation task~\cite{ChenZXQZ19, LeeCLS19, TangW18, ZhangXZZ20}. However, these studies focus on the conventional recommendation task and are not suitable for conversational systems, where users progressively express their personalized preferences over conversational turns. This means that these studies do not account for the interactive way of user feedback over conversational turns.

\section{Conclusion}\label{sec:con}
This study demonstrated the imperative need to downsize the high training cost of conversation recommendation systems. Researchers should take into account the number of parameters when designing a neural architecture for conversational recommendation systems. One possible way to combat this issue is to follow a knowledge distillation strategy. Nevertheless, the key challenge is to consider the continuous user-item interactions over the conversational turns when following a knowledge distillation strategy.

\balance
\bibliographystyle{./bibliography/IEEEtran}
\bibliography{./bibliography/IEEEabrv,./bibliography/IEEEexample}

\end{document}